\documentclass{article}

\usepackage{microtype}
\usepackage{graphicx}
\usepackage{booktabs} % for professional tables

\usepackage[utf8]{inputenc} % allow utf-8 input
\usepackage[T1]{fontenc}    % use 8-bit T1 fonts
\usepackage{hyperref}       % hyperlinks
\usepackage{url}            % simple URL typesetting
\usepackage{booktabs}       % professional-quality tables
\usepackage{amsfonts}       % blackboard math symbols
\usepackage{nicefrac}       % compact symbols for 1/2, etc.
\usepackage{microtype}      % microtypography
\usepackage{wrapfig}
\usepackage{graphicx} % more modern
\usepackage{epsfig} % less modern

\usepackage{color}

\usepackage{float}
\usepackage{caption}
\usepackage{subcaption}

\usepackage{amsmath,amssymb,bm} % define this before the line numbering.
\usepackage{amssymb}
\usepackage{amsmath}
\usepackage{amsthm}
\usepackage{mathtools}
\usepackage[raggedright]{titlesec} % avoid hyphen in sections

\usepackage{hyperref}

\DeclareMathOperator{\Tr}{Tr}

\def\vec[#1]{\mbox{\boldmath $#1$}}
\def \mat     [#1]{\mathbf {#1}}
\def \matEqSub[#1 #2]{\mathbf{#1}_{#2}}
\def \matEquation[#1]{{\mathbf{#1}}}
\def \matCell[#1]{\mbox{$\bf #1_{cell}$}}
\def \mtilda [#1]{\mathbf{\widetilde{#1}}}

\newcommand{\matrixdim}[2] {#1 \times #2}

\usepackage[accepted]{icml2018}

\icmltitlerunning{Low-rank geometric mean metric learning}

\begin{document}

\onecolumn
\icmltitle{Low-rank geometric mean metric learning}



\icmlsetsymbol{equal}{*}

\begin{icmlauthorlist}
\icmlauthor{Mukul Bhutani}{iisc}
\icmlauthor{Pratik Jawanpuria}{msft}
\icmlauthor{Hiroyuki Kasai}{tok}
\icmlauthor{Bamdev Mishra}{msft}
%\icmlauthor{Aeiau Zzzz}{equal,to}
%\icmlauthor{Bauiu C.~Yyyy}{equal,to,goo}
%\icmlauthor{Cieua Vvvvv}{goo}
%\icmlauthor{Iaesut Saoeu}{ed}
%\icmlauthor{Fiuea Rrrr}{to}
%\icmlauthor{Tateu H.~Yasehe}{ed,to,goo}
%\icmlauthor{Aaoeu Iasoh}{goo}
%\icmlauthor{Buiui Eueu}{ed}
%\icmlauthor{Aeuia Zzzz}{ed}
%\icmlauthor{Bieea C.~Yyyy}{to,goo}
%\icmlauthor{Teoau Xxxx}{ed}
%\icmlauthor{Eee Pppp}{ed}
\end{icmlauthorlist}

\icmlaffiliation{iisc}{Indian Institute of Science <mukul.bhutani93@gmail.com>.}
\icmlaffiliation{msft}{Microsoft, India <\{pratik.jawanpuria,bamdevm\}@microsoft.com>.}
\icmlaffiliation{tok}{The University of Electro-Communications, Japan <kasai@is.uec.ac.jp>}

\icmlcorrespondingauthor{Mukul Bhutani}{mukul.bhutani93@gmail.com}
\icmlcorrespondingauthor{Bamdev Mishra}{bamdevm@microsoft.com}

\icmlkeywords{Metric learning, Grassmann, Manopt}

\vskip 0.3in



\printAffiliationsAndNotice{}  % leave blank if no need to mention equal contribution

\begin{abstract}
We propose a low-rank approach to learning a Mahalanobis metric from data. Inspired by the recent geometric mean metric learning (GMML) algorithm, we propose a low-rank variant of the algorithm. This allows to jointly learn a low-dimensional subspace where the data reside and the Mahalanobis metric that appropriately fits the data. Our results show that we compete effectively with GMML at lower ranks.
\end{abstract}

\section{Introduction}
We propose a scalable solution for the {\it Mahalanobis} metric learning problem \cite{Kulis}. The Mahalanobis distance is defined as $d_{\mat[A]}(\vec[x],\vec[x']) = (\vec[x]-\vec[x'])^{\top}\mat[A](\vec[x]-\vec[x'])$, where $\vec[x], \vec[x'] \in \mathbb{R}^{d}$ are input vectors and $\mat[A]$ is a $\matrixdim{d}{d}$ symmetric positive definite (SPD) matrix. The objective is to learn a suitable SPD matrix $\mat[A]$ from the given data. Since $\mat[A]$ is a $d\times d$ SPD matrix, most state-of-the-art metric learning algorithms scale poorly with the number of features $d$~\cite{Harandi}. To mitigate this, a pre-processing step of dimensionality reduction (e.g., by PCA) is generally applied before using popular algorithms like LMNN and ITML \cite{Weinberger2009,Davis2007}.

Recently, \cite{Zadeh2016} proposed the geometric mean metric learning (GMML) formulation, which enjoys a closed-form solution. However, it requires matrix $\mat[A]$ to be positive definite, which makes it unscalable in a high dimensional setting. To alleviate this concern, we propose a low-rank decomposition of $\mat[A]$ in the GMML setting. Low-rank constraint also has a natural interpretation in the metric learning setting, since the group of similar points in the given dataset reside in a low-dimensional subspace. We jointly learn the low-dimensional subspace along with the metric. We show that the optimization is on the Grassmann manifold and propose a computationally efficient algorithm. On real-world datasets, we achieve competitive results comparable with the GMML algorithm, even though we work on a smaller dimensional space. 

\section{Problem formulation}
We follow a weekly supervised approach in which we are provided two sets $\mathcal{S}$ and $\mathcal{D}$ containing pairs of input points belonging to same and different classes respectively. Taking inspiration from GMML, we formulate the objective function as:
\begin{equation}\label{eq:full_formulation}
\begin{array}{ll}
\min\limits_{\mat[A] \succeq 0} & \Tr(\mat[AS]) + \Tr(\mat[A^{\dagger}D])\\
{\rm subject\ to} & {\rm rank}(\mat[A]) = r,
\end{array}
 \end{equation}
where $\mat[S]:= \sum_{(\vec[x]_i , \vec[x]_j) \in \mathcal{S}}{(\vec[x]_i-\vec[x]_j)(\vec[x]_i-\vec[x]_j)^{\top}}$, $\mat[D] := \sum_{(\vec[x]_i , \vec[x]_j) \in \mathcal{D}}{(\vec[x]_i - \vec[x]_j)(\vec[x]_i - \vec[x]_j)^{\top}}$, and $\mat[A^{\dagger}]$ is the pseudoinverse of $\mat[A]$.

Exploiting a particular fixed-rank factorization \cite{Meyer2010}, we factorize rank-$r$ matrix $\mat[A]$ as $\mat[A] = \mat[UBU]^\top$, where $\mat[U]$ is an orthonormal matrix of size $d\times r$ and $\mat[B] \succ 0$ is of size $r\times r$. Consequently, we rewrite (\ref{eq:full_formulation}) as:
\begin{equation}
\label{eq:rank_constrained_formulation}
\min\limits_{\mat[U]^\top \mat[U] = \mat[I]}\ \min\limits_{\mat[B] \succ 0} \quad \Tr(\mat[UBU^{\top}S]) + \Tr(\mat[U]\mat[B]^{-1}\mat[U]^{\top}\mat[D]).
\end{equation}

If we define $\mtilda[S] = \mat[U^{\top}SU]$ and $\mtilda[D] = \mat[U^{\top}DU]$, then the inner minimization problem with respect to $\mat[B]$ has a closed-form solution as the geometric mean of ${\mtilda[S]}^{-1}$ and $\mtilda[D]$ \cite{Zadeh2016}, i.e.,  $\mat[B] = \mtilda[S]^{-1/2} (\mtilda[S]^{1/2} \mtilda[D] \mtilda[S]^{1/2})^{1/2}   \mtilda[S]^{-1/2}$. Using this fact, the outer optimization problem is readily checked to be only on the column space of $\mat[U]$. To see this, consider the group action $\mat[U] \mapsto \mat[UO]$, where $\mat[O]$ is an $r$-by-$r$ orthogonal matrix. For all $\mat[O]$, this group action captures the column space of $\mat[U]$. The mapping $\mat[U] \mapsto \mat[UO]$ implies that $\mat[B] \mapsto \mat[O]^\top \mat[BO]$ and consequently, the objective function in (\ref{eq:rank_constrained_formulation}) remains invariant to all $\mat[O]$. 

The set of column spaces is the abstract Grassmann manifold ${\rm Grass}(r,d)$, which is defined as the set of $r$-dimensional subspaces in $\mathbb{R}^d$. We represent the abstract column space of $\mat[U]$ as $\mathcal{U}$. Equivalently, (\ref{eq:rank_constrained_formulation}) is formulated as:
\begin{equation}
\label{eq:grassmann_formulation}
\begin{array}{lll}
\min\limits_{\mathcal{U}\in {\rm Grass}(r,d)} & f(\mathcal{U}),
\end{array}
\end{equation}
where $f(\mathcal{U}):= \min\limits_{\mat[B]\succ 0} \Tr(\mat[UBU^{\top}S]) + \Tr(\mat[U]\mat[B]^{-1}\mat[U]^{\top}\mat[D])$. Optimization on the Grassmann manifold has been a well-researched topic in the literature \cite{absil08a}. Optimization algorithms are developed conceptually with abstract column spaces $\mathcal{U}$, but by necessity are implemented with matrices $\mat[U]$.

Extending the idea to a setting which weighs the sets $\mathcal{S}$ and $\mathcal{D}$ unequally, we obtain the formulation
\begin{equation}
\label{eq:weighted_rank_constrained_formulation}
\begin{array}{lll}
\min\limits_{\mat[U]^\top \mat[U] = \mat[I]}\ \min\limits_{\mat[B] \succ 0}  \quad (1 - t) \delta_{R}^2(\mat[B],(\mat[U^{\top}SU])^{-1}) + \ t \delta_{R}^2(\mat[B], \mat[U^{\top}DU]),
\end{array}
\end{equation}
where $\delta_{R}$ denotes the Riemannian distance on the SPD manifold and $t \in[0,1]$ is a hyperparameter. Similarly to (\ref{eq:rank_constrained_formulation}), the problem (\ref{eq:weighted_rank_constrained_formulation}) is also on the Grassmann manifold as the inner problem has a closed-form solution as the weighted geometric mean between ${\mtilda[S]}^{-1}$ and $\mtilda[D]$. For $t=0.5$, (\ref{eq:weighted_rank_constrained_formulation}) is equivalent to (\ref{eq:rank_constrained_formulation}).

\section{Implementation}
Our proposed algorithm LR-GMML is implemented using the Matlab toolbox Manopt \cite{boumal14a} that provides an interface to develop optimization algorithms on the Grassmann manifold for (\ref{eq:grassmann_formulation}). Specifically, we use the conjugate gradients solver. It only requires the matrix expression of the objective function and the partial derivative of the objective function with respect to $\mat[U]$ to be supplied. Particularly, for (\ref{eq:rank_constrained_formulation}), the expression is 
\begin{equation}\label{eq:partial_derivatives}
\begin{array}{ll}
\text{the cost function wrt }\mat[U] = &(1 - t) \delta_{R}^2(\mat[B],(\mat[U^{\top}SU])^{-1}) + \ t \delta_{R}^2(\mat[B], \mat[U^{\top}DU])\\
\text{the partial derivative wrt }\mat[U] = &(1-t)\ \mat[SU] \mtilda[S]^{-1/2} {\rm logm}(\mtilda[S]^{1/2} \mat[B] \mtilda[S]^{1/2}) \mtilda[S]^{-1/2}\\
&\quad  - t\ \mat[DU] \mtilda[D]^{-1/2} {\rm logm}(\mtilda[D]^{-1/2} \mat[B] \mtilda[D]^{-1/2}) \mtilda[D]^{-1/2},
\end{array}
\end{equation}
where ${\rm logm}$ is the matrix logarithm of a matrix, $\mtilda[S] = \mat[U^{\top}SU]$, $\mtilda[D] = \mat[U^{\top}DU]$, and $\mat[B] = \mtilda[S]^{-1/2} (\mtilda[S]^{1/2} \mtilda[D] \mtilda[S]^{1/2})^t   \mtilda[S]^{-1/2}$. The Riemannian distance $\delta_R^2(\mat[B],\mat[Y])$ has the matrix expression $\|{\rm logm} (\mat[B]^{-1/2} \mat[Y] \mat[B]^{-1/2})\|_F^2$. The Grassmann specific requirements are handled internally by the Manopt toolbox.

The computational cost of implementing the conjugate gradients solver for (\ref{eq:weighted_rank_constrained_formulation}) depends on two set of ingredients: the objective function specific ingredients and the Grassmann specific ingredients. The computational cost of the objective function related ingredients depend on computing $\mtilda[S]$ and $\mtilda[D]$, which cost $O(ndr + nr^2)$, where $n$ is the total number of pairs of the points. Computing $\mat[B]$ and partial derivative in (\ref{eq:partial_derivatives}) cost $O(r^3)$. Computing the Grassmann specific ingredients cost $O(dr^2)$ \cite{boumal14a,absil08a}. Overall, the computational cost of our algorithm is linear in $n$ and avoids quadratic complexity on $d$. In comparison, the GMML algorithm of \citet{Zadeh2016} costs $O(d^3)$.
 
The code is available at \url{https://github.com/muk343/LR-GMML}. 

\section{Results} 
\begin{figure}[H]
%\vskip -0.2in
\begin{center}
\centerline{\includegraphics[width=\columnwidth]{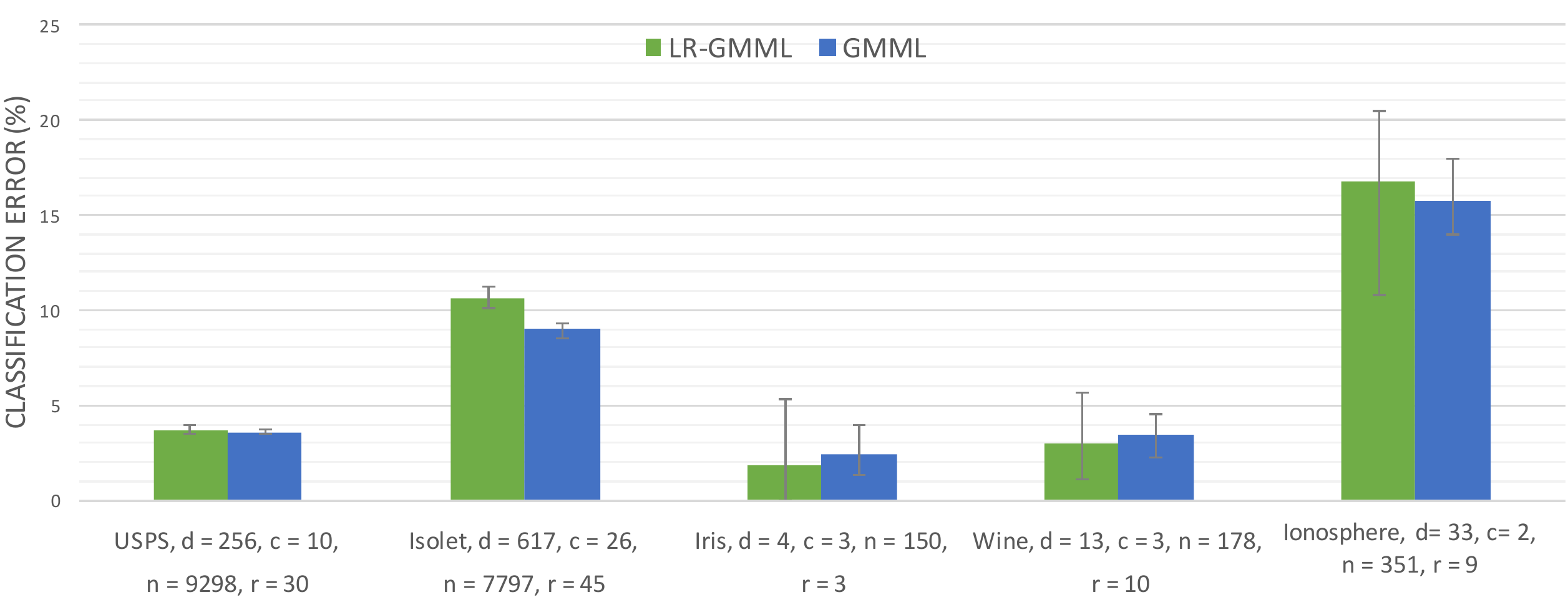}}
\caption{Classification error rates of k-NN classifier comparing LR-GMML with GMML. Dimensionality of feature space ($d$), number of classes ($c$), number of total data points ($n$), and the rank used in LR-GMML ($r$) are listed below each dataset.}
\label{fig:error}
\end{center}
\end{figure}

\begin{figure}[H]
    \begin{subfigure}[H]{0.50\textwidth}
    \centering
        \includegraphics[width=0.99\linewidth]{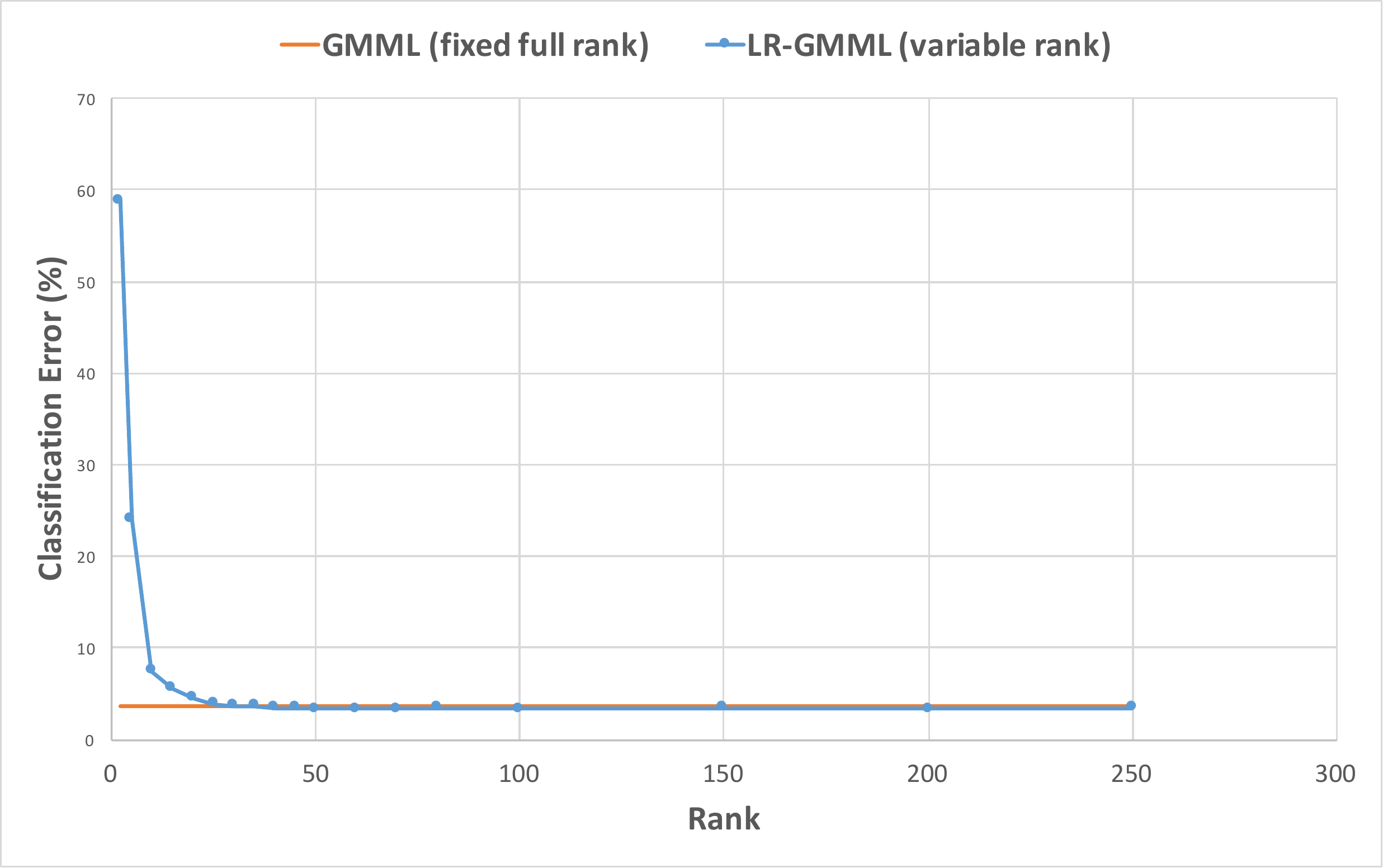}
        \caption{USPS}
        \label{fig:dcostU}
    \end{subfigure}\hfill%
    \begin{subfigure}[H]{0.50\textwidth}
    \centering
        \includegraphics[width=0.99\linewidth]{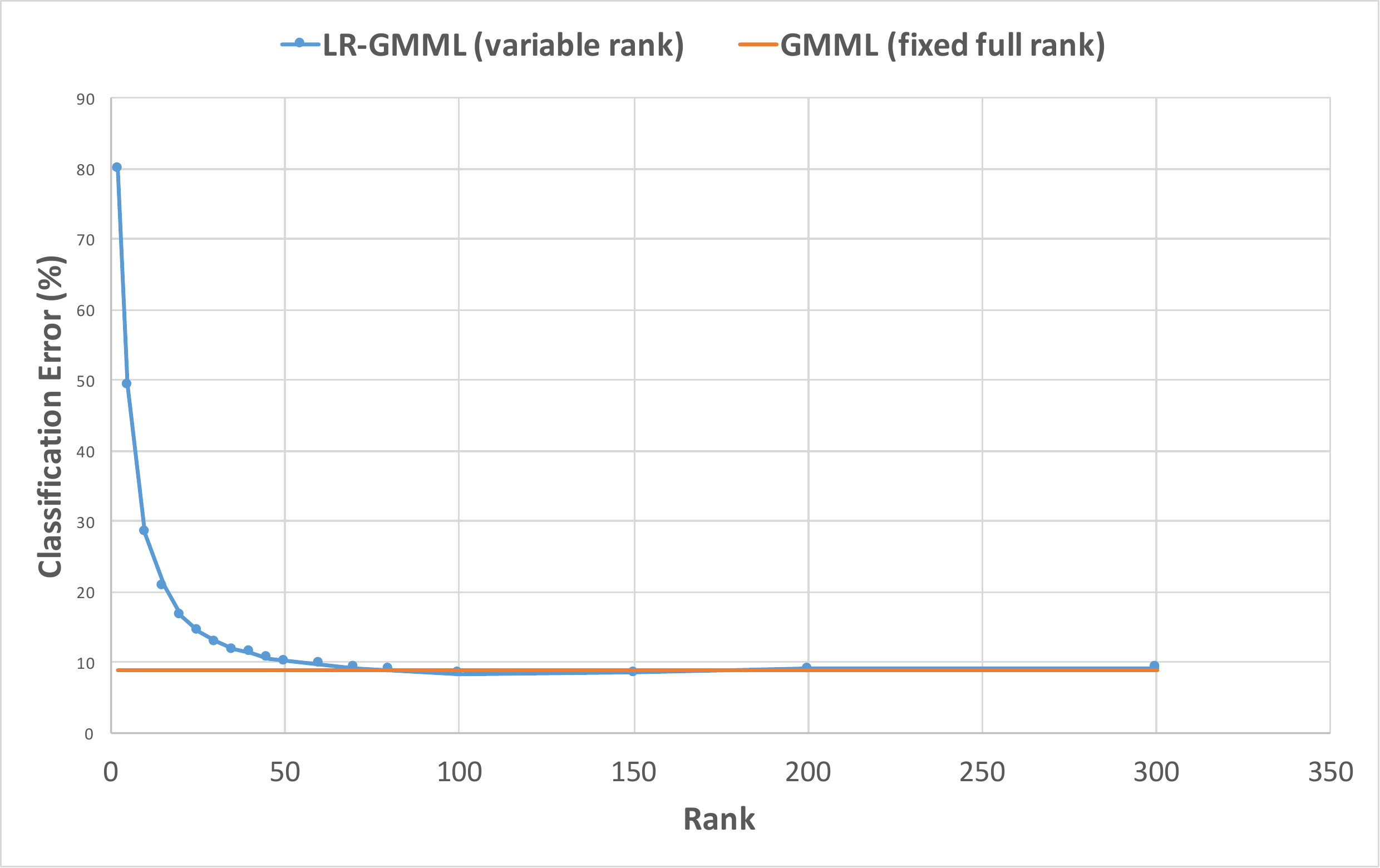}
        \caption{Isolet}
        \label{fig:dcostW}
    \end{subfigure}\hfill%
    \caption{Classification error rates for LR-GMML for various low rank settings in comparison with that of GMML.} 
    \label{fig:variableRank}
\end{figure}

We compare LR-GMML with GMML on publicly available UCI datasets by measuring the classification error for a k-NN classifier following the procedure in \cite{Zadeh2016}. Parameter $t$ is optimized for both the algorithms and average errors over five random runs are reported in Figure \ref{fig:error}. It can be seen that our proposed method is able to get results comparable to GMML even while dealing with low rank. Figure \ref{fig:variableRank} delineates this comparison by showing the accuracy LR-GMML achieves for various ranks (shown in blue) in comparison to GMML (shown in orange). Here also, we use similar settings and report accuracy numbers averaged over five random runs.

%Our algorithm is performing as good as GMML in most of the datasets even while working with low rank. This difference in rank is even more apparent with large datasets like USPS and Isolet. \\ Leveraging Riemannian geometry, we presented a low rank metric learning approach which was able to perform as well as current state of the art full rank approach.  

\section{Conclusion}
We presented a low-rank variant of the geometric mean metric learning (GMML) algorithm. The problem is shown to be an optimization problem on the Grassmann manifold, for which we exploit the Riemannian framework to develop a conjugate gradients solver. Our results show that even at lower-ranks, our low-rank variant LR-GMML performs as good as the full space (standard) GMML algorithm on many standard datasets. In future, we intend to work on larger datasets as well as in an online learning setting. Another direction is exploring LR-GMML with recently proposed deep learning techniques.

\bibliography{metric_learning_icml_arXiv}
\bibliographystyle{icml2018}

\end{document}